\documentclass[preprint,10pt,3p]{elsarticle}
\usepackage{graphics} 
\usepackage{epsfig} 
\usepackage{times} 
\usepackage{amsmath} 
\usepackage{amssymb}  
\usepackage{epstopdf}
\usepackage{subfigure,color,balance}
\usepackage{caption}
\usepackage{verbatim}
\usepackage{cases}
\allowdisplaybreaks[4]
\usepackage{enumerate}
\usepackage{booktabs}
\usepackage{balance}
\usepackage{mathbbol}
\usepackage{algorithm}
\usepackage{algorithmicx}
\usepackage{algpseudocode}
\usepackage{mathrsfs}
\usepackage{threeparttable}
\usepackage[draft]{todonotes}
\usepackage{multirow}
\usepackage{makecell}

\usepackage{enumitem}
\usepackage{bm}

\newcommand{\method}[1]{\texttt{#1}}

\usepackage[colorlinks=true,      
linkcolor=black,      
citecolor=black,      
filecolor=black,      
urlcolor=blue]{hyperref}

\newdefinition{remark}{Remark}

\def\x{{\bf x}}

\def\0{{\bf 0}}
\def\1{{\bf 1}}

 \biboptions{compress}

\begin{document}
	
	\begin{frontmatter}

		\title{{S2O: An Integrated Driving Decision-making Performance Evaluation Method\\ Bridging Subjective Feeling to Objective Evaluation}}
		
		
		\author[a]{Yuning Wang}
		\ead{wangyn20@mails.tsinghua.edu.cn}
		\author[b]{Zehong Ke}
		\ead{kzh21@mails.tsinghua.edu.cn}
		\author[a]{Yanbo Jiang}
		\ead{jyb23@mails.tsinghua.edu.cn}
		\author[a]{Jinhao Li}
		\ead{qingxu@tsinghua.edu.cn}
		\author[a]{Shaobing Xu}
		\ead{shaobxu@tsinghua.edu.cn}
		\author[c]{John M. Dolan\corref{cor1}}
		\ead{jdolan@andrew.cmu.edu}
        \author[a]{Jianqiang Wang\corref{cor1}}
		\ead{wjqlws@tsinghua.edu.cn}
		
		\cortext[cor1]{Corresponding author: John M. Dolan and Jianqiang Wang}
		

		\address[a]{School of Vehicle and Mobility, Tsinghua University, 100084 Beijing, China}
		\address[b]{Xingjian College, Tsinghua University, 100084 Beijing, China}
        \address[c]{Robotics Institute, Carnegie Mellon University, Pittsburgh, PA 15213 USA}
		
\begin{abstract}
Autonomous driving decision-making is one of the critical modules towards intelligent transportation systems, and how to evaluate the driving performance comprehensively and precisely is a crucial challenge. A biased evaluation misleads and hinders decision-making modification and development. Current planning evaluation metrics include deviation from the real driver trajectory and objective driving experience indicators. The former category does not necessarily indicate good driving performance since human drivers also make errors and has been proven to be ineffective in interactive close-loop systems. On the other hand, existing objective driving experience models only consider limited factors, lacking comprehensiveness. And the integration mechanism of various factors relies on intuitive experience, lacking precision. In this research, we propose \method{S2O}, a novel integrated decision-making evaluation method bridging subjective human feeling to objective evaluation. First, modified fundamental models of four kinds of driving factors which are safety, time efficiency, comfort, and energy efficiency are established to cover common driving factors. Then based on the analysis of human rating distribution regularity, a segmental linear fitting model in conjunction with a complementary SVM segment classifier is designed to express human’s subjective rating by objective driving factor terms. Experiments are conducted on the D2E dataset, which includes approximately 1,000 driving cases and 40,000 human rating scores. Results show that \method{S2O} achieves a mean absolute error of 4.58 to ground truth under a percentage scale. Compared with baselines, the evaluation error is reduced by 32.55\%. Implementation on the SUMO platform proves the real-time efficiency of online evaluation, and validation on performance evaluation of three autonomous driving planning algorithms proves the feasibility.

%
%

\end{abstract}
		
		
	\end{frontmatter}
	
\section{Introduction}

Autonomous Driving (AD) is expected to reshape the future transportation system, significantly improving traffic efficiency and safety. Following the SAE automation-level definition, many industrial manufacturers have proposed L2+ AD applications, making vehicles autonomous in regular traffic scenes. However, it is still challenging for current AD technologies to cope with complicated interactive scenarios, such as intersections mixed with various kinds of agents, crowded highway merges, etc.~\cite{wang2023decision,tang2022prediction} 

AD can be divided into submodules including perception, situation awareness, decision and planning, control, etc. With the breakthroughs of deep learning technologies, some studies have used end-to-end frameworks so that some submodules can be merged. Nevertheless, whatever technical approach is used, a planning result always needs to be output and executed on the vehicle end. Decision-making under complex traffic environments is jointly influenced by drivers, vehicles, and the environment~\cite{xiong2023integrated}, thus making it one of the most important and difficult challenges toward high-level AD.

\begin{figure*}[tb!]
	\centering
	\includegraphics[width=10cm]{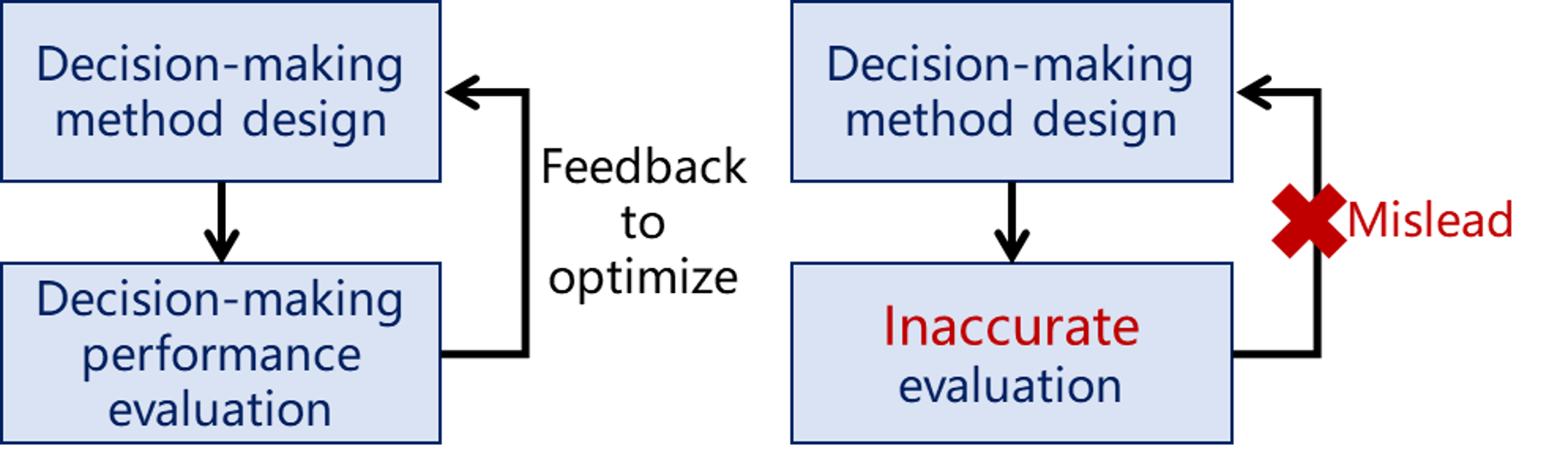}
	\caption{The impact of inaccurate evaluation on decision-making development.}  
	\label{Fig:impact of evaluation}
\end{figure*}

Fig.~\ref{Fig:impact of evaluation} demonstrates the development process of AD decision-making methods. After a decision-making algorithm is designed, performance evaluation is conducted so that the algorithm can be iteratively revised and optimized based on the test results, forming a closed loop~\cite{zhan2018probabilistic}. However, if the evaluation results are biased or inaccurate, it will mislead and hinder the algorithm development, forming a vicious cycle. For instance, in the NuPlan AD planning challenge, the final evaluation indicator is the simple sum of several physical quantities~\cite{caesar2021nuplan}, which made the algorithm design deviate from the original intention. In order to improve the final score, some planning algorithms tended to make decisions contrary to common human behavior such as over-conservative deceleration, irregular frequent lane changing, etc.~\cite{dauner2023parting}, and the fundamental reason lies in the invalidity of evaluation methods and metrics. Therefore, reasonable decision performance evaluation methods and indicators are crucial for AD decision-making algorithm development. 

Different from perception or control problems which have clear physical definitions of the ground truth and metrics such as recognition accuracy, static control error, etc., decision-making performance evaluation is a subjective and complex process conducted by humans~\cite{kumar2024prediction}. Currently there is no universally accepted or standard metric for AD decision-making, and how to appropriately design an evaluation model is still an open challenge~\cite{zhan2018probabilistic}.

Existing decision-making evaluation methods can be divided into two categories. The first category considers the vehicle's deviation from the real trajectories of human drivers and is widely used in data-driven planning methods. In recent years, many large-scale datasets aimed at AD planning have emerged~\cite{wang2024survey}, involving Nuscene, Waymo Open Motion, HighD, etc. Based on these datasets, many studies used the L2 distance error from the real future trajectories or the degree of anthropomorphism to evaluate the performance of their proposed planning models~\cite{caesar2020nuscenes,ettinger2021large,wang2022high}. However, even though some dataset publishers claimed that their drivers were experienced and carefully selected, it is still inappropriate to regard human driving results as the ground truth. Firstly, the actual trajectory is merely one solution among numerous feasible trajectories and may not necessarily represent the optimal one. Even the most experienced drivers also make mistakes, and the ultimate goal of AD is to surpass humans rather than to match them~\cite{wang2021towards}. On the other hand, this offline open-loop evaluation metric is difficult to adapt to real dynamic closed-loop driving scenarios. As a continuous process, the decision in the previous time-step will influence the surrounding agents’ behaviors in the future, resulting in error accumulation effects~\cite{xu2024sequence}. According to Phong’s study, imitation learning methods are even inferior to the constant-velocity planning in closed-loop driving simulations~\cite{phong2024truly}. Hence, static trajectory distance errors cannot assess the driving performance over a period of time.

In addition to fitting to real driver trajectories, in recent years some studies have attempted to evaluate decision-making algorithms from the perspective of objective driving experience~\cite{zhan2018probabilistic}. Li et al. assessed the driving comfort performance by calculating the average vibration level~\cite{li2016road}. Sohrabi et al. defined the relative crash rate as a safety effectiveness indicator~\cite{sohrabi2021quantifying}. Wang et al. analyzed the travelling time, vehicle pose, and accident rates to determine whether the proposed decision-making algorithm was better than the baselines~\cite{wang2023differentiated}. Although evaluating planning performance in an objective manner, these studies mostly focused on only single or a few specific driving elements and lacked comprehensiveness. According to vehicle theory, driving experience should contain four essential factors: time efficiency, energy consumption, safety, and comfort~\cite{gillespie2021fundamentals, li2023intelligent}. Therefore, only considering limited aspects of driving experience will result in a biased evaluation.

Some studies explored integrated methods to consider multiple driving factors. Huang et al. developed an integrated driving evaluation system based on the principle of least action quantity which is the sum of risk energy and vehicle virtual kinetic energy~\cite{huang2020integrated}. Zhao et al. used the CRITIC and AHP algorithms to derive the relative weights of vehicle physical quantities~\cite{zhao2022objective}. Some recent AD planning challenges such as Nuplan, Carla leaderboard, onsite, etc. also ranked competitors by a comprehensive driving performance score including safety, comfort, efficiency, etc.. The scores are derived by directly adding up single factor terms~\cite{caesar2021nuplan}. Although these studies have formed a framework for comprehensive evaluation, the modeling of individual factors is simplistic. Furthermore, the relative relationships or weights between different factors primarily rely on naive experience, failing to express human's generalized subjective feelings. There is still significant potential for improving the accuracy and comprehensiveness of driving decision-making performance evaluation.

The core challenge in decision evaluation is how to use objective physical models to reflect the subjective assessment process from humans. On one hand, objective driving performance evaluation contains four objective performance aspects: safety, comfort, time efficiency, and energy consumption. On the other hand, during the human's subjective evaluation process, instead of giving rating on individual experience terms, people collectively consider these four aspects of factors and directly generate an overall subjective evaluation result. In this process, the integration mechanism of various elements and their respective weights are unclear, posing a challenge in establishing evaluation models.

Toward these critical challenges, in this study we propose S2O, a novel integrated decision-making performance evaluation method bridging subjective human feeling to objective evaluation.  By combining and revising previous research, we first establish the single-term evaluation models for safety, time efficiency, comfort, and energy consumption. Subsequently, based on the collected human-vehicle-environment dataset D2E~\cite{ke2024d2e}, we design a comprehensive AD decision-making performance evaluation model composed of segmented linear fitting, term normalization, segment classification, and collision penalty revision, learning integration regularity from human evaluation intelligence. By fitting to subjective evaluation data, S2O derives the relative weights of various objective driving factors. Experiments and validations are conducted on both offline datasets and online evaluation systems. The main contributions of this study are listed below:
\begin{itemize}
	\item Establish modified single-term driving performance evaluation models of safety, time efficiency, comfort, and energy consumption containing varieties of driving factors.
	\item Propose an evaluation integration model based on segmented linear fitting, SVM segment classification, normalization, and crash revision to learn relative factor weights from human subjective evaluation data.
	\item Experiment results prove that compared with baselines, the proposed method significantly decreases the evaluation error from human ground truth.
\end{itemize}

The following sections of this paper are organized as follow: Section 2 reviews the related works about driving factor modelling and integration methods. Section 3 introduces the framework and details of S2O method. Experiment results, analysis, and validations are demonstrated in Section 4. Section 5 gives the conclusions and future works.

\section{Related works}
\label{Sec:2}

\subsection{Driving Experience Factor Modeling}

Some previous studies have proposed single-term decision-making evaluation models to assess planning performance. In this research, we combine their strengths and establish comprehensive single-term evaluation models.

Safety is a commonly used evaluation metric in decision-making studies, and collision judgement is the most intuitive method. He et al. recorded the average collision time to denote the accident frequency~\cite{he2022robust}. Sobrabi et al. further divided collisions into minor and severe according to the intensity~\cite{sohrabi2021quantifying}. Discrete crash counting is meaningful but not elaborate, since those cases where no collision occurs can also be dangerous. Therefore, continuous indicators are now widely used. Weng et al. used the minimum time-to-collision (TTC) and time headway (THW) to illustrate the level of danger~\cite{weng2020model,ortiz2023road}. Tejeda et al. defined a crash probability based on a finite-state machine system~\cite{tejada2019towards}. Although these metrics can evaluate safety continuously, most methods are scenario-driven and lack scene generalization. Recently, field-based methods, which generate a two-dimensional energy field to represent the risk distribution, have gained more and more attention~\cite{zhang2023importance}. Wang et al. proposed a Driving Safety Field (DSF) to calculate the risk to the ego vehicle from other agents and the environment~\cite{wang2015driving}. Because of the continuity and scene generalization, this study establishes the safety evaluation model based on modified DSF.

As for time efficiency, the main considered factors are velocity speed and total time consumed to finish a task, which are actually equivalent to each other~\cite{wang2023decision,chen2021mixed}. Note that the road speed limit should be regarded as a constraint, and faster is not always better. Excessive speed should be penalized. Comparatively, a model based on the ratio of vehicle speed and speed limit combined with an overspeed penalty is more comprehensive than simply calculating average section velocity~\cite{wang2020research}.

Comfort is also an important aspect of the driving experience, since carsickness is one of the most common reasons that cause unpleasant riding. The main contributors to discomfort include vibrations, sudden intense moves, frequent large turns, etc.~\cite{iskander2019car,zhang2024overtaking}. Nha et al. defined a comfort penalty metric by counting the number of lane changes~\cite{nha2012comparative}. McAllister et al. calculate the accumulated square sum of longitudinal and lateral jerk during a driving event~\cite{mcallister2017concrete}. Li et al. evaluated the square root of velocity vibrations~\cite{li2016road}. In this study, we combine these factors into an integrated model.

Energy consumption has been more and more emphasized given the interest in green transportation. The first type of calculation method focuses on the energy resources of the power unit, which is fuel consumption for gasoline vehicles and electricity consumption for electric vehicles~\cite{fiori2016power}. Another type of method computes the energy consumed by the vehicle end, which is composed of four terms: acceleration, wind resistance, road gradient resistance, and rolling friction~\cite{gillespie2021fundamentals}. In this study, to eliminate the impact of vehicle brand and power unit efficiency differences, we use the latter method as the energy consumption calculation model.

\subsection{Evaluation Integration Methods}

The core challenge of decision-making evaluation is to integrate single assessment terms into one comprehensive model. Existing studies can be divided into three categories: direct sum model, correlation analysis, and regression fitting.

Current evaluation models considering multiple driving experience factors mainly use the direct sum to give the final assessment result. Both Nuplan and the Onsite AD planning challenge listed a series of performance-related indicators based on physical quantities and summed them up with a collision revision multiplier~\cite{caesar2021nuplan,zhi2023construction}. Based on a direct summing-up model, Wang et al. added a maximum truncation threshold for each term~\cite{wang2020research}. Carla leaderboard further refined the value of the revision multiplier for collision with various agent types~\cite{dosovitskiy2017carla}. While the goal of merging multiple individual terms into one model can be achieved, direct summing-up is not accurate because various factor terms have difference value domains and levels of discrimination. Some factors may lose their effectiveness, and the final score is dominated by the term with the highest gradient.

Correlation analysis is widely used in the field of social sciences, and some scholars have applied it to decision-making evaluation. Rehm et al. assessed driving performance by forming a virtual knowledge graph, referring to expert experience to create the evaluation topology~\cite{rehm2024virtual}. Yin et al. used a decision tree model to give a final score for a driving event~\cite{yin2017multi}. Zhao et al. used criteria importance through the inter-criteria correlation method to determine the relative importance of various physical factors and analyzed the similarity to expert drivers by grey relational analysis~\cite{zhao2022objective}. Compared with the direct sum approach, these methods are more elaborate and can reflect relative importance. However, the derived correlation coefficients are not equivalent to the needed weights in the evaluation process, and such methods rely on expert experience to obtain open-loop hyperparameters which are not precise enough.

In other evaluation fields such as economy and statistics, regression models show good accuracy and have interpretability. Wang et al. used linear fitting to merge safety, comfort, and efficiency into one model~\cite{wang2023novel}. Based on vanilla linear fitting, improved versions are developed including locally weighted regression which sets a bandwidth for each data point~\cite{cleveland1988locally}, multivariate adaptive regression, which fits the data by a set of primary functions~\cite{zhou2007predicting}, and segmental linear fitting which separates the segments according to data distribution features and uses different weights for each one. Due to the fitting procedure, regression-based methods have higher accuracy while maintaining interpretability, and the main challenge lies in obtaining suitable datasets. In this study, based on the established human-vehicle-environment dataset D2E~\cite{ke2024d2e}, we utilize segmental linear fitting together with revision multiplier as the backbone integration model.

\section{Methods}
\subsection{Framework}

\begin{figure*}[tb!]
	\centering
	\includegraphics[width=16cm]{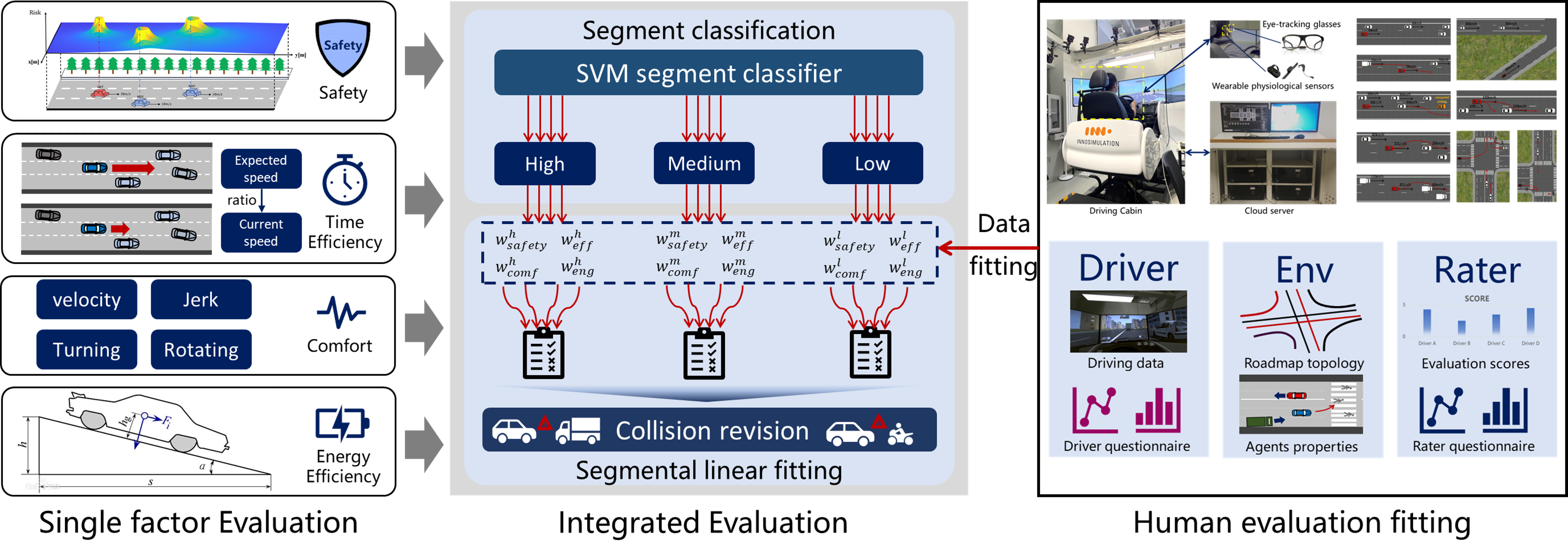}
	\caption{An illustration of S2O decision evaluation framework.}  
	\label{Fig:framework}
\end{figure*}

The framework of the proposed S2O AD decision-making performance evaluation method is shown in Fig.\ref{Fig:framework}. First four fundamental models of driving experience, which are safety, time efficiency, comfort, and energy consumption, are established to give a comprehensive single-term evaluation. With the input of four computed term scores which are normalized to percentage scales, an integrated evaluation model is designed to derive the final assessment result of driving events. The SVM classifier first judges the case level class, then a segment linear model is applied to output the holistic evaluation score. Finally, a collision revision module is utilized to give larger penalties to accidents, ensuring the priority of human driving safety. The core hyperparameters in the model, which are the weights of four factors, are derived through fitting to subjective human evaluation data collected in the D2E dataset. S2O reflects the subjective evaluation patterns of humans through objective physical assessment models, achieving a comprehensive quantitative evaluation of driving decision-making and planning events.

\subsection{Single Factor Evaluation Models}

To derive a comprehensive evaluation method, four objective evaluation methods should be established first, which are safety, time efficiency, comfort, and energy efficiency.

\subsubsection{Safety evaluation model}

As introduced in the literature review, a risk energy field is selected to assess safety because of its continuity, generalization ability, and comprehensiveness. Based on the study of Wang et al.~\cite{wang2015driving}, we revise the Driving Safety Model (DSF) to fit the demands of safety evaluation.

\begin{figure*}
    \centering
    \includegraphics[width=10cm]{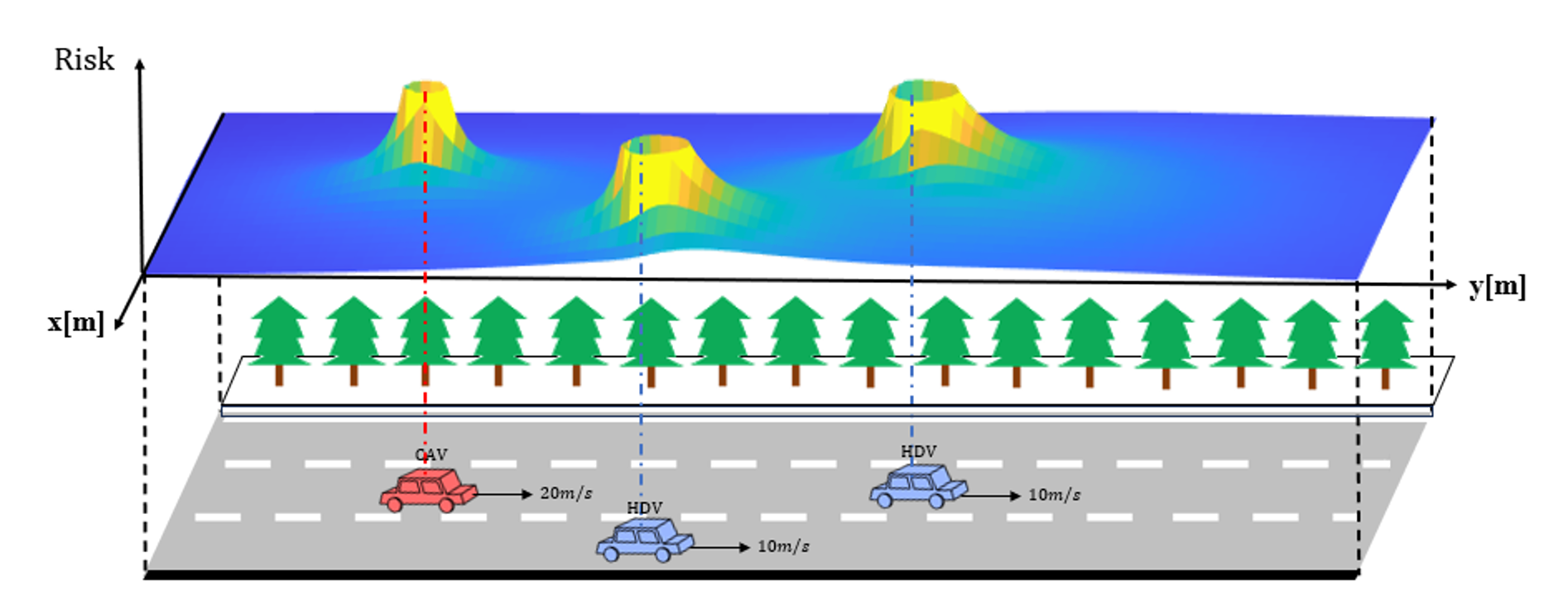}
    \caption{An illustration diagram of Driving Safety Field}
    \label{Fig: DSF}
\end{figure*}

The basic idea of DSF is to treat each agent on the road as a risk source radiating risk energy. The risk a vehicle is exposed to is the sum of the risk energy generated by all agents at that location, as shown in Fig.~\ref{Fig: DSF}. In this study, we improve the definition of the safety energy model by considering the concept of virtual mass and equivalent distance.

The safety risk a vehicle at position ($x$,$y$) has from an agent at position ($x_a$,$y_a$ ) is defined as~\eqref{static energy}:

\begin{equation}\label{static energy}
    R_a = \frac{G M_{eq}}{r_{eq}^2} + \frac{k_1 \exp(v_r \cos \theta) }{r_{eq}^2 }
\end{equation}

\noindent where $M_{eq}$ is the virtual mass, $r_{eq}$ is the equivalent distance, $v_r$ is the relative velocity value, and $\theta$ is the angle between the relative velocity vector and the distance vector. $k_1$ and $G$ are constant coefficients.

The virtual mass $M_{eq}$ is the revised mass of a vehicle, since a vehicle with higher velocity will cause more severe collision results. The expression of  $M_eq$ is shown in~\eqref{equal mass}:

\begin{equation}
    \label{equal mass}
    M_{eq} = M (a\cdot v^b_{a\_ long} + c)
\end{equation}

\noindent where $M$ is the mass of the risk source agent, $v^b_{a\_ long}$ is the agent’s longitudinal velocity, and $a$, $b$, and $c$ are constant parameters.

The equivalent distance $r_{eq}$ is a revision of the vanilla distance calculation considering the motion sensitivity of the ego vehicle in different directions. The definition is shown in~\eqref{equal distance}:

\begin{equation}
    \label{equal distance}
    r_{eq} = \sqrt{r_{long}^2 + k_a\cdot r_{lat}^2}
\end{equation}

\noindent where $r_{long}$ and $r_{lat}$ represent the longitudinal and lateral distance between the ego vehicle and the risk agent respectively, and $k_a$ is the scaling factor of the agent. In this study, $k_a$ is the aspect ratio of the bounding box size. By weighting on the longitudinal and lateral directions, the risk area of the agents is no longer circular but elliptical, better reflecting the differences in longitudinal and lateral risk.

$R_{ego}$, the overall risk of the ego vehicle at position $(x,y)$  is the superposition of the risks generated by all agents in the scene at position $(x,y)$, as expressed in~\eqref{overall risk} where $A$ is the agent set. And the final safety term score $S_{safety}$ is the average risk energy intensity during driving, as shown in~\eqref{safety evaluation} where $t_f$ and $t_s$ are the finishing and starting time.

\begin{equation}
    \label{overall risk}
    R_{ego} = \sum_{a \in A} R_a
\end{equation}

\begin{equation}
    \label{safety evaluation}
    S_{safety} = \frac{\int_{t_s}^{t_f} R_{ego}dt}{t_f - t_s}
\end{equation}

Note that when agents are too far away from the ego vehicle, the risk energy they generate can be considered negligible compared to those that are closer. To improve computational efficiency, in this study, we consider an area of interest (ROI) of 100 meters to the front and 50 meters to the rear. In~\eqref{safety evaluation}, only agents within the ROI are taken into calculation.

\subsubsection{Time efficiency evaluation model}
Time efficiency is usually measured by the ratio of the current velocity and expected speed which in most cases is the road speed limit. When speed is slow, the quicker the better. However, penalties should be given for excessive speeding. Taking China’s traffic law as an example, speeding less than 20\% can be tolerated, while more than that will be seen as a violation. In addition, exceeding the speed limit by over 50\% is considered as severe speeding. In our time efficiency evaluation design, we follow these regulations, as shown in~\eqref{efficiency def}:

\begin{equation}
    \label{efficiency def}
    s_{eff} = \begin{cases}
        1 - \frac{v}{v_{lim}}& {v < v_{lim}}\\
        0& {v_{lim} < v \leq 1.2 v_{lim}} \\
        \min (1, \frac{10v}{3v_{lim} - 4})& {v > 1.2 v_{lim}}
    \end{cases}
\end{equation}

\noindent where $v$ is the vehicle velocity $v_{lim}$ is the speed limit. The values of $v_{lim}$ under various scenarios are shown in Table~\ref{Tb:roadSpeedLimit}. Under the designed model, time efficiency evaluation maintains consistency with the safety term such that higher values indicate poorer driving performance. Excessive speeding penalty begins when $v$ is larger than 1.2 $v_{lim}$ and reaches maximum penalty when $v$ is larger than 1.5 $v_{lim}$.

\begin{table}[htb]
\footnotesize
	\begin{center}
		\caption{Speed limits of various road types}\label{Tb:roadSpeedLimit}
		\begin{threeparttable}
		\setlength{\tabcolsep}{5mm}{
		\begin{tabular}{cccccccc}
		\toprule
			Road section type & Speed limit $(km/h)$\\\hline
			 Urban regular road & $60$\\
    Urban intersection & 30\\
    Highway slow lane & 80\\
    Highway express lane & 120 \\
			\bottomrule
		\end{tabular}}
		\end{threeparttable}
	\end{center}
	\vspace{-5mm}
\end{table}

The final evaluation score of time efficiency is the average value of $s_{eff}$ during the driving event, as illustrated in~\eqref{efficiency evaluation}.

\begin{equation}
    \label{efficiency evaluation}
    S_{eff} = \frac{\int_{t_s}^{t_f}s_{eff}dt}{t_f - t_s}    
\end{equation}

\subsubsection{Comfort evaluation model}
As introduced in related works, basic comfort evaluation components involve jerk, trajectory curvature, etc. Meanwhile, velocity also influences the discomfort level for passengers while riding, since centripetal force experienced by people is proportional to the product of the vehicle's angular velocity and linear speed.

Apart from these vehicle state-related quantities, specific behaviors such as repetitive U-turns and emergency stops also significantly decrease the comfort level. Hence, in this study we add a discrete term to punish the unpleasant motions.

The overall comfort model is described in~\eqref{comfort def}, where $\omega$ is the angular velocity, $v$ is the linear velocity, $j$ is the jerk, $L_{upm}$ is the loss caused by unpleasant motions which is a behavior counter, and  $k$ is a constant weighting coefficient. Similar to other factors, the final comfort score of a driving case $S_{comf}$ is the temporal weighted average value of $s_{comf}$, as shown in~\eqref{comfort evaluation}.

\begin{equation}
    \label{comfort def}
    s_{comf} = \omega \cdot v + k \cdot j^2 + L_{upm}
\end{equation}

\begin{equation}
    \label{comfort evaluation}
    S_{comf} = \frac{\int_{t_s}^{t_f}s_{comf}dt}{t_f - t_s}
\end{equation}

\subsubsection{Energy efficiency evaluation model}
In our study, we use the total energy consumed for driving as the indicator to evaluate the energy efficiency so that differences on powertrain systems, which are irrelevant to the decision-making algorithms, will not influence the assessment results. According to vehicle dynamic theories, the total consumed energy can be decomposed into four parts: acceleration, wind resistance, road gradient resistance, and rolling friction~\cite{gillespie2021fundamentals}.

The acceleration term represents the energy used for velocity increasing. The calculation is given by~\eqref{P_j}, where $\delta$ is the conversion coefficient of the rotating mass, $m$ is the vehicle mass, $u_a$ (km/h) is the vehicle speed, and $u$ (m/s) is also the vehicle speed.

\begin{equation}
    \label{P_j}
    P_j = \frac{\delta m u_a}{3600} \cdot \frac{du}{dt}
\end{equation}

As for the wind resistance, the calculation is shown in~\eqref{P_w}, where $C_D$ is the air resistance coefficient, and $A$ is the windward area of the car.

\begin{equation}
    \label{P_w}
    P_w = \frac{C_D A u_a^3}{76140}
\end{equation}

Road gradient resistance can be computed by~\eqref{p_i}, where $g$ is gravitational acceleration and $i$ is the road gradient.

\begin{equation}
    \label{p_i}
    P_i = \frac{mgiu_a}{3600}
\end{equation}

Road rolling friction can be derived by~\eqref{P_f}, where f is the road surface coefficient.
\begin{equation}
    \label{P_f}
    P_f = \frac{Gfu_a}{3600}
\end{equation}

The total consumed energy is the sum of~\eqref{P_j} to~\eqref{P_f}, as shown in~\eqref{p sum}. And the final energy term score of the whole driving $S_{eng}$ event is computed by~\eqref{energy evaluation}.

\begin{equation}
    \label{p sum}
    P_e = P_j+P_w+P_i+P_f
\end{equation}

\begin{equation}
    \label{energy evaluation}
    S_{eng} = \frac{\int_{t_s}^{t_f}P_e dt}{t_f - t_s}
\end{equation}

\subsection{Evaluation Integration Model}
\label{3.C}

After deriving the single-term assessment results, an integration model is needed to evaluate driving cases comprehensively. In the literature review we surveyed related integration methods, and to increase evaluation accuracy while maintaining interpretability, a revised segmental linear model is selected, including normalization, segment classifier, and integration considering crash revision.

\subsubsection{Normalization}
One critical problem is that the magnitudes of the four single-term scores are different, ranging from single digits to hundreds. However, what matters is the relative variance instead of absolute values. Therefore, in this study we first normalized all terms into an identical range, which is $[60,100]$.

Common normalization methods include max-min linear normalization, $arctan$ curve, robust normalization, z-score, etc. After comparison, the result of which will be shown in the experiments section, we selected max-min linear normalization as the normalization style. Max-min linear normalization is a common method of data normalization that transforms data into a unified value range. The standardized value of term $i$ $S_i^{std}$ can be calculated by~\eqref{normalization}, where $S_i$ is the original value of the term, $S_i^{min}$ is the minimum value of the data, and $S_i^{max}$ is the maximum value. The value ranges from 60 to 100.

\begin{equation}
    \label{normalization}
    S_i^{std} = \frac{S_i - S_i^{min}}{S_i^{max} - S_i^{min}} \cdot 40 + 60
\end{equation}

\subsubsection{Segment classification based on SVM}
Decision-making evaluation is not a simple linear problem. Based on the analysis of the human rater evaluation score distribution, it can be observed that for different levels of driving events, people pay varying degrees of attention to each factor.

\begin{figure*}
    \centering
    \includegraphics[width=10cm]{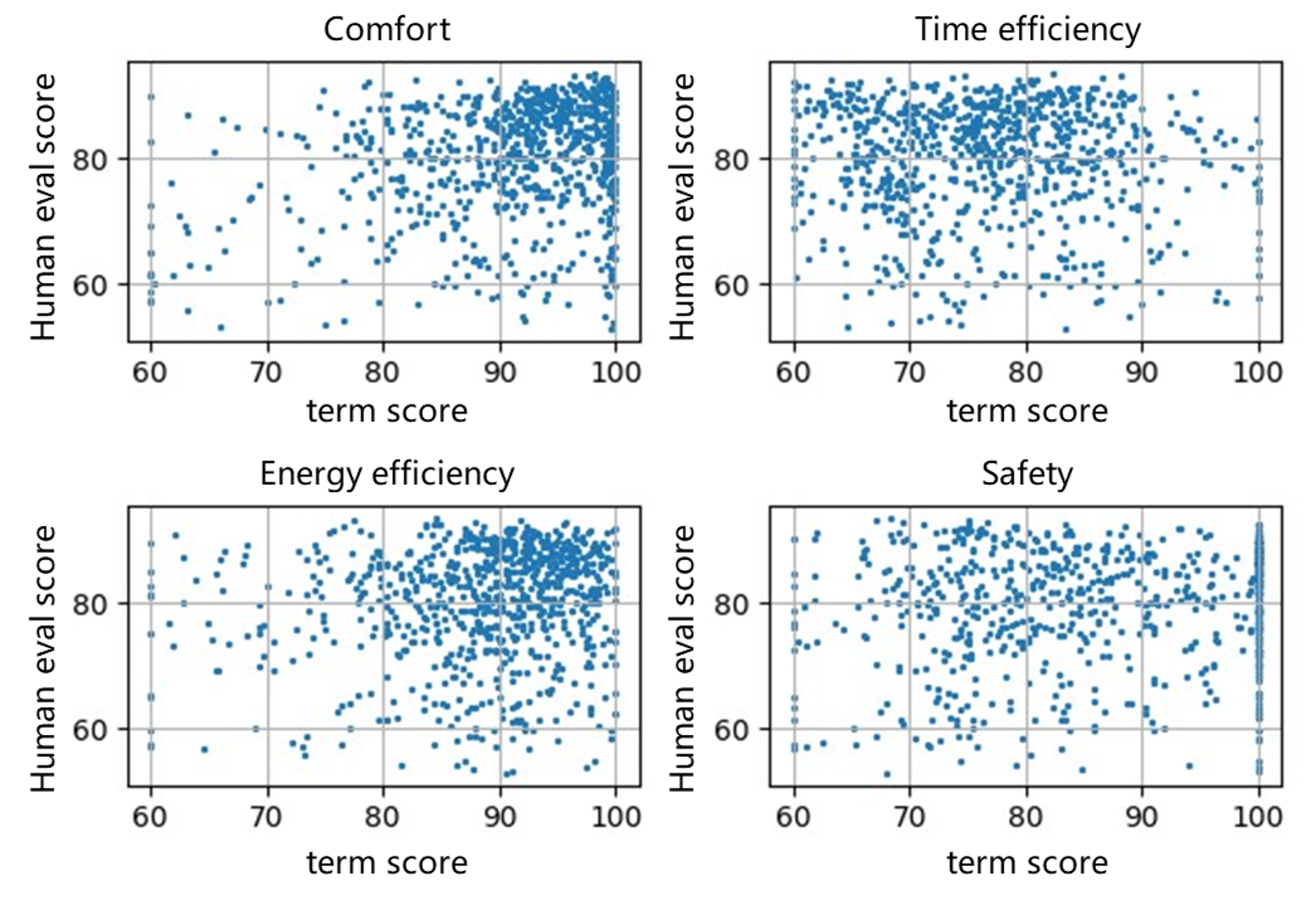}
    \caption{Distribution between single factors and human evaluation results}
    \label{Fig: distribution}
\end{figure*}

Fig.~\ref{Fig: distribution} demonstrates the distribution between the overall evaluation ground given by human raters in the D2E dataset and four normalized single-term scores. Each point represents a driving event, with the horizontal axis representing the corresponding term score and the vertical axis representing the rater's score. The original human score range is $[0,100]$. According to human rating criteria in the D2E dataset, 60 is considered as the borderline of acceptable driving, and events with extremely low scores are often associated with collisions and other incidents, leading to anomalies in the values of the four individual metrics. To eliminate the disturbances caused by abnormal events and observe the patterns in regular driving evaluations, we exclude all events involving collisions from the distribution. Hence, the lower bound of human rating scores for the remaining events is close to 60. In the later integrated evaluation method design, we will use a revision multiplier to restore the impact of collisions.

According to the holistic analysis of the distribution, we can observe that no single element exhibits a significant linear or simple monotonic relationship with human scoring, which further proves that driving decision-making evaluation is a complicated process influenced by a multitude of factors and cannot be simplistically represented by a single term. The distribution results demonstrate the necessity of an integrated evaluation method considering multiple factors.

On the other hand, from the perspective of each factor, we can discover that humans have varying factor sensitivity to driving events with various score levels. Generally speaking, comfort is strongly correlated with the high score range. Only when the comfort level is high can the case achieve a good overall score. Comparatively, the energy efficiency factor has good discrimination ability on all driving events with scores above 60. The safety term, with a wider distribution, is more likely a prerequisite of good driving, thus having little impact on the high score range. However, for the mid and low range safety generates many differences. Similarly, the time efficiency factor shows great impact on the mid and low range, while making little difference in the high range. Considering the fact that humans’ evaluation mechanisms of various driving event levels are different, in this study we separate driving evaluation into three segments and fit a set of weights for each one to reflect diverse human sensitivity towards driving events of different levels.

\begin{figure*}
    \centering
    \includegraphics[width=8cm]{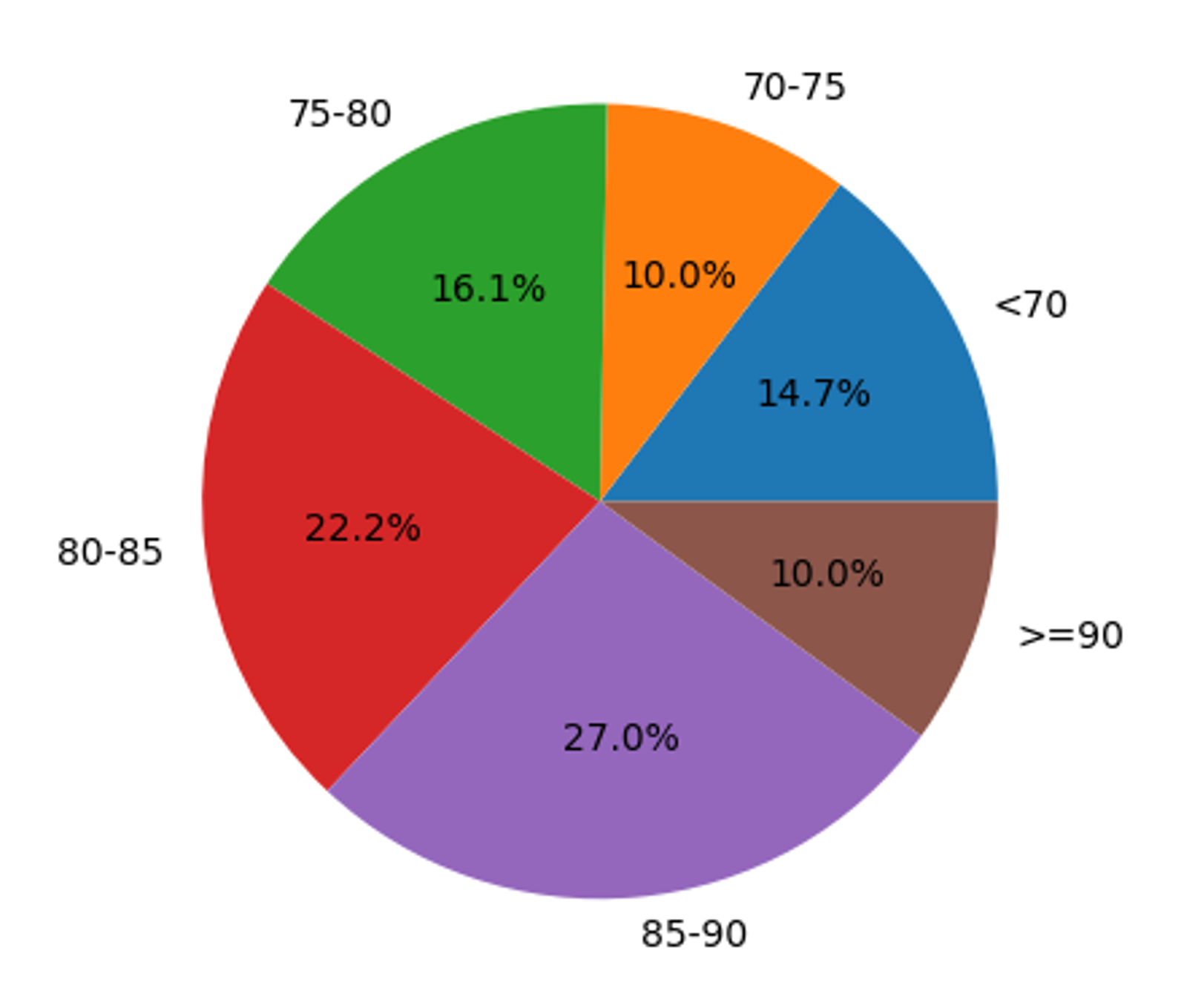}
    \caption{Frequency distribution of human's evaluation score}
    \label{Fig: score distribution}
\end{figure*}

Fig.~\ref{Fig: score distribution} shows the frequency of human evaluation scores. According to the distribution, we divide the evaluation into three intervals: low: [0, 75], mid: (75, 85], and high: (85, 100], accounting for 24.7\%, 38.3\%, and 37.0\% of the full set of evaluation scores respectively. This segmentation also aligns with human intuitive impressions of high, medium, and low scores.

Since in the later linear fitting module, we will use different weights for various segments, it is necessary to determine which interval an event falls into first, which is a classification problem. To balance the interpretability and accuracy, we use Support Vector Machine (SVM) as the classifier model. The primary objective of SVM is to find the hyperplane that best separates the data points into different classes while maximizing the margin. The decision boundary or hyperplane is determined by a linear combination of the input features, represented by ~\eqref{SVM} where $\bm{w}$ is the weights and $b_c$ is the bias term. $\bm{x}$ is the input vector, which is the combination of the four single-term evaluation scores in this study.

\begin{equation}
    \label{SVM}
    f(x) = \bm{w^T x} + b_c
\end{equation}

The goal of SVM is to find the optimal hyperplane that maximizes the margin, which can be formulated as the following optimization problem~\eqref{SVM class} where $y_i$ is the segment label of driving events, $N$ is the total number of data points.

\begin{equation}
    \label{SVM class}
    \begin{aligned}
    &\min{\frac{1}{2}{||\bm{w}||}^2} \\
    &s.t. y_i(\bm{w^T x} + b_c) \geq 1, for i=1,...,N  
    \end{aligned}
\end{equation}

Apart from the SVM, a pre-screening for collision is conducted to filter the crashed cases. Considering that safety should always be the priority of driving, once a collision happens the classifier will label the driving event as the lowest level.

After training the SVM classifier, the segmental linear fitting can be realized to precisely assess driving decisions.

\subsubsection{Segmental linear fitting}
With the four normalized single factor scores and the segment classification, a segmental linear fitting model is applied to give the integrated decision-making evaluation score, as shown in~\eqref{segmental fitting}, where $w_{safe}$, $w_{eff}$, $w_{comf}$, and $w_{eng}$ are the corresponding weights of the terms, and $b$ is the constant offset. The offset is used to compensate for the static offset in human ratings. For instance, in many events, raters assign extremely high scores exceeding 95 points. However, it’s hard for objective algorithms to achieve scores close to the maximum. Therefore, the offset helps to decrease the error.

\begin{equation}
    \label{segmental fitting}
    S_{int} = w_{safe}S_{safe} + w_{eff}S_{eff} +w_{comf}S_{comf}+w_{eng}S_{eng}+b
\end{equation}

As a regression problem, the weights of three segments are derived by fitting on the D2E dataset, the process and results of which will be introduced in the experiment section.

Apart from the four mentioned terms above, the human evaluation of driving experience also includes a veto mechanism for crashes. 

Although the safety term can partly reflect the risk when a crash happens, since it is an average value on the whole driving case, more severe punishment should be given when sudden extreme danger happens. Otherwise, even when an accident occurs, the score can still be tens of points, which does not meet the human’s priority for absolute safety. Referring to the Nuplan simulation~\cite{caesar2021nuplan} and Carla leaderboard~\cite{dosovitskiy2017carla}, crash revision can be expressed by a punishment multiplier as shown in~\eqref{crash def}. The final revised integrated evaluation score $S_{int}^{rev}$ is given by~\eqref{crash revise}.

\begin{equation}
    \label{crash def}
    C = \begin{cases}
        1& {if crash}\\
        0& {no crash} \\
    \end{cases}
\end{equation}

\begin{equation}
    \label{crash revise}
    S_{int}^{rev} = C \cdot S_{int}
\end{equation}

\section{Implementation, Experiments, and Analysis}
\label{Sec:4}

In this study, we use D2E, a dataset involving driving scenario data and human evaluation scores of driving events, to implement the proposed S2O method and compare the evaluation performances. First, the hyperparameters including the weights of difference segments are derived by fitting to human evaluation ground truth. Comparisons with several baselines including both single-term models and comprehensive ones are given to analyze the performance differences. In addition, feasibility is validated through the implementation of the proposed S2O evaluation framework on SUMO platform, conducting real-time driving decision-making evaluation. We also verified the effectiveness of the proposed S2O by evaluating three AD planning algorithms.

\subsection{Dataset Selection}
We use D2E as the dataset to implement and validate the performance of the proposed S2O AD decision-making evaluation framework. D2E is a dataset collected by our team previously, encompassing closed-loop data collection of the entire driving process involving human-vehicle-environment evaluation. Detailed information about D2E can be found in~\cite{ke2024d2e}.

\begin{figure*}
    \centering
    \includegraphics[width=10cm]{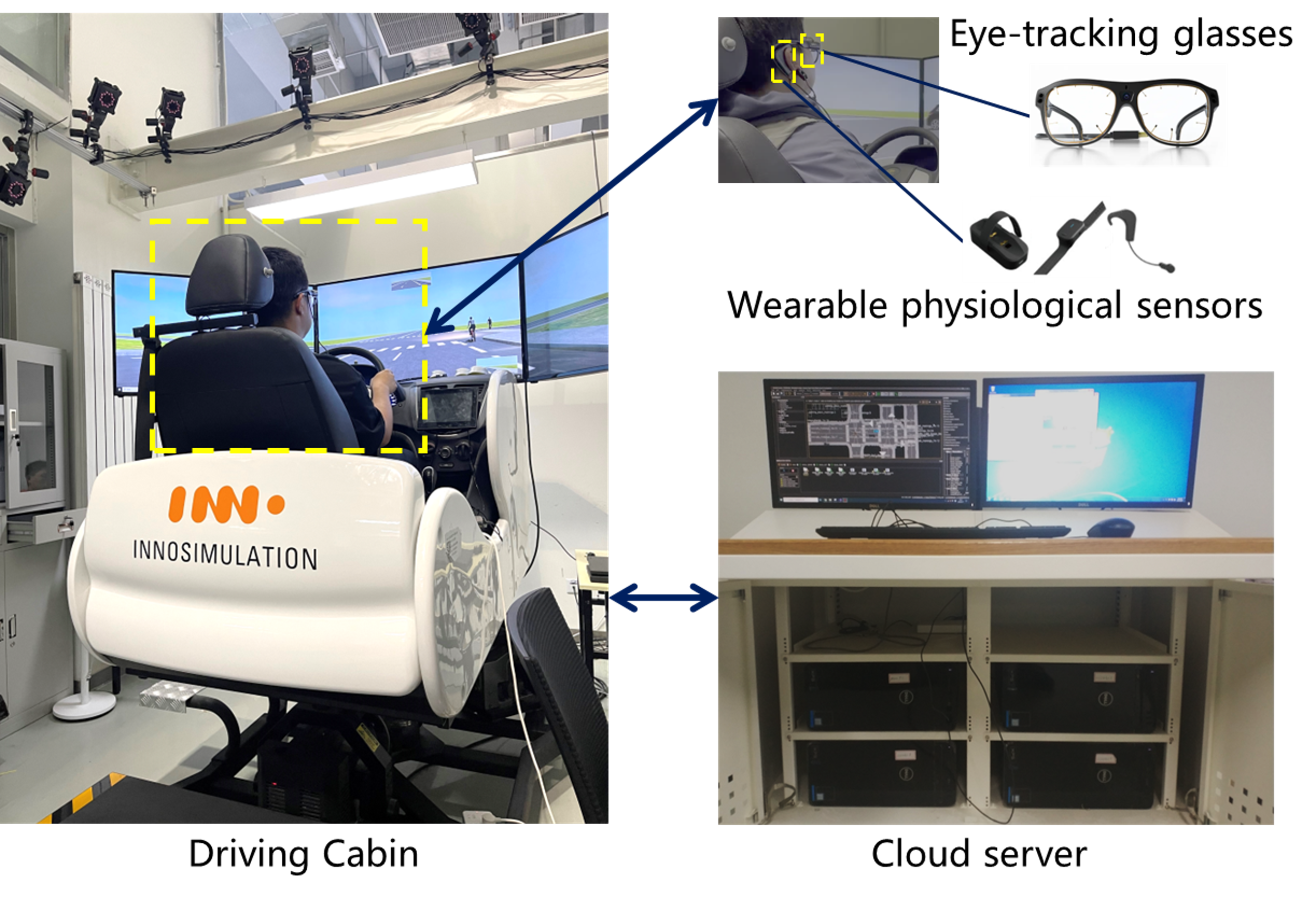}
    \caption{Data collection platform}
    \label{Fig: data platform}
\end{figure*}

In D2E, referring to the category of other public datasets such as Waymo Open Motion~\cite{ettinger2021large}, Nuplan~\cite{caesar2021nuplan}, Argoverse~\cite{chang2019argoverse}, etc., 12 kinds of interactive scenes are designed in a driving simulator platform, including unprotected intersections, highway ramp, crowded urban road, etc. The data are collected through a driving simulator composed of the driving cabin with a motion platform, eye-tracking glasses, driver wearable physiological sensors, and cloud server, as shown in Fig.~\ref{Fig: data platform}.

\begin{figure*}
    \centering
    \includegraphics[width=10cm]{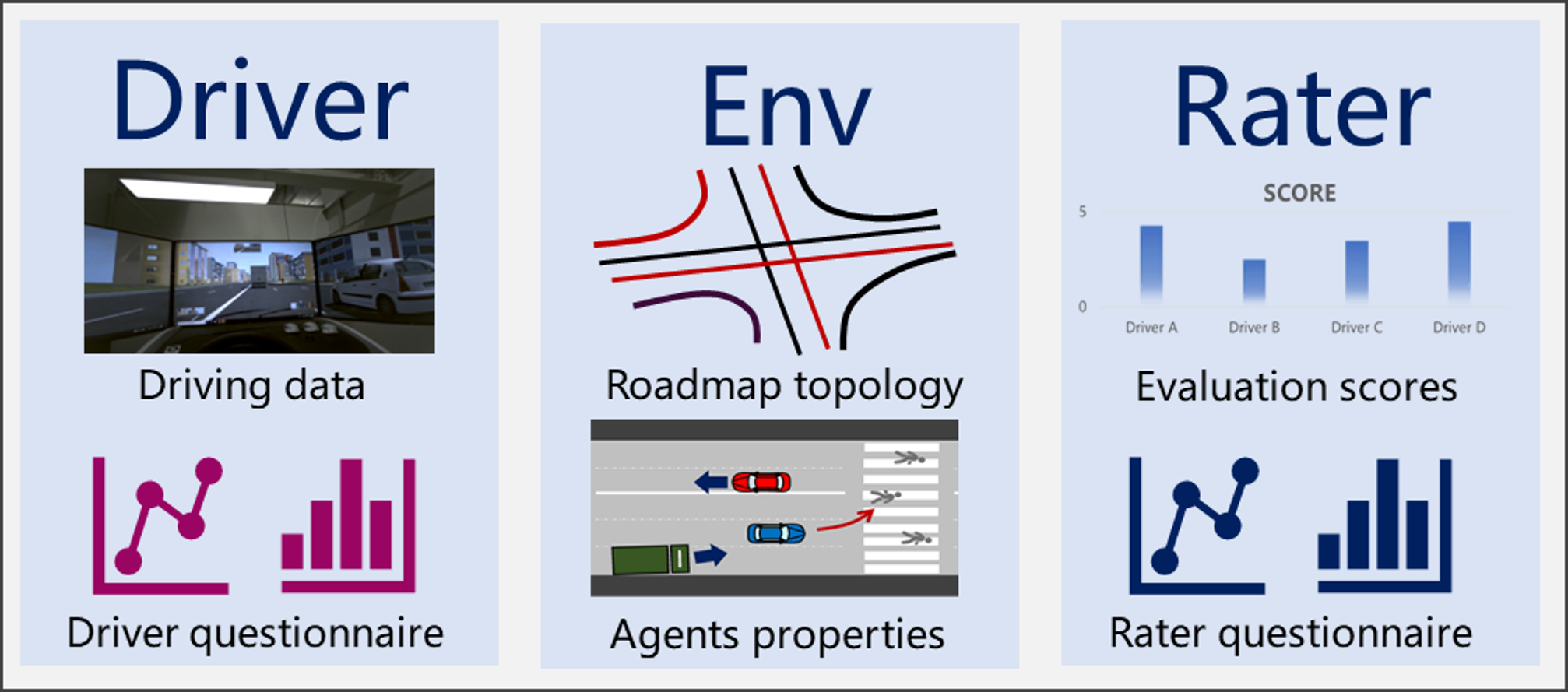}
    \caption{Data used for decision-making performance evaluation}
    \label{Fig: data type}
\end{figure*}

The data types involved in D2E and used in this experiment are illustrated in Fig.~\ref{Fig: data type}. On the driver end, the driving manipulation data are recorded, and personal information containing historical driving experience, driving style, etc. are collected by questionnaire. As for the driving environment data, the simulator platform stores all needed data for calculating the objective evaluation terms involving roadmap topology and agents’ properties such as positions, velocities, sizes, etc. In addition, third-party volunteers are recruited as raters to give subjective scores after watching first-person-view recorded videos collected by the eye-tracking glasses. Evaluation scores are on a percentage scale, with 60 points considered as the acceptable borderline and 90 points considered as the borderline for excellent driving. Information also includes rater preferences on various aspects of the riding experience as recorded in the questionnaires.

The volume of data in D2E is sufficient to support our research. 80 drivers drove 12 scenarios, with each event of each driver scored by 40 raters, resulting in a total of 38,400 data labels. The distributions of drivers and raters are wide, including various genders, driving ages, driving and preference styles, etc., as shown in Table.~\ref{Tb:driver distribution} and Table.~\ref{Tb:rater distribution}. For raters, since not only drivers but also passengers have the right to evaluate driving events, 14 volunteers who do not have a driving license were also recruited to make the assessment distribution more comprehensive.

\begin{table}[htb]
\footnotesize
	\begin{center}
		\caption{Driver distribution of the dataset}\label{Tb:driver distribution}
		\begin{threeparttable}
		\setlength{\tabcolsep}{5mm}{
		\begin{tabular}{cccccccc}
		\toprule
			Genre & Feature & Number & Ratio\\\hline
			 \multirow{2}{*}{Gender} & Male & 68 & 85.0\%\\
                                        & Female & 12 & 15.0\% \\
                \multirow{4}{*}{Driving years} & 0-3&15&18.8\%\\
                                                   & 4-9 & 30 & 37.5\% \\
                                                   &10-19 & 28 & 35\% \\
                                                    &20+ & 7 &8.8\% \\
                \multirow{3}{*}{Driving style} & Conservative & 35 & 43.8\%\\
                                                    &Medium & 36 & 45.0\% \\
                                                    &Aggressive & 9 & 11.3\% \\
                \multirow{3}{*}{Accident experience} & None & 53 & 66.3\% \\
                                                    & Minor accidents & 20 & 25.0\% \\
                                                    & Moderate accidents & 7 & 8.8\% \\
			\bottomrule
		\end{tabular}}
		\end{threeparttable}
	\end{center}
	\vspace{-5mm}
\end{table}

\begin{table}[htb]
\footnotesize
	\begin{center}
		\caption{Rater distribution of the dataset}\label{Tb:rater distribution}
		\begin{threeparttable}
		\setlength{\tabcolsep}{5mm}{
		\begin{tabular}{cccccccc}
		\toprule
			Genre & Feature & Number & Ratio\\\hline
			 \multirow{2}{*}{Gender} & Male & 20 & 50.0\%\\
                                        & Female & 20 & 50.0\% \\
                \multirow{2}{*}{Driving license} & Have & 26 & 65.0\%\\
                                                   & Don't have & 14 & 35.0\% \\
                \multirow{2}{*}{Preference style} & Conservative & 18 & 45.0\%\\
                                                    &Medium & 11 & 27.5\% \\
                                                    &Aggressive & 11 & 27.5\% \\
			\bottomrule
		\end{tabular}}
		\end{threeparttable}
	\end{center}
	\vspace{-5mm}
\end{table}

To ensure the data quality, we use several preprocessing steps to eliminate the invalid data. The first step is to filter the driving cases that are too short to give an appropriate evaluation. In these cases, drivers made mistakes such as driving on curbs or crashing into other agents shortly after the scene began. After filtering, 943 driving events are preserved. The second step is to eliminate the invalid rating scores. We found that some raters just gave identical scores to all events, even though some of them experienced collisions. Therefore, we remove those raters whose rating variance is lower than 5 and those who have a difference of more than 30 points from the others’ average. The last step is to calculate the ground truth of the driving cases. For each event, we compute the mean evaluation score after excluding the top 10\% and the bottom 10\% of rating scores, serving as the fitting target of the objective evaluation model.

\subsection{Model Implementation and Fitting}

\begin{figure*}
    \centering
    \includegraphics[width=8cm]{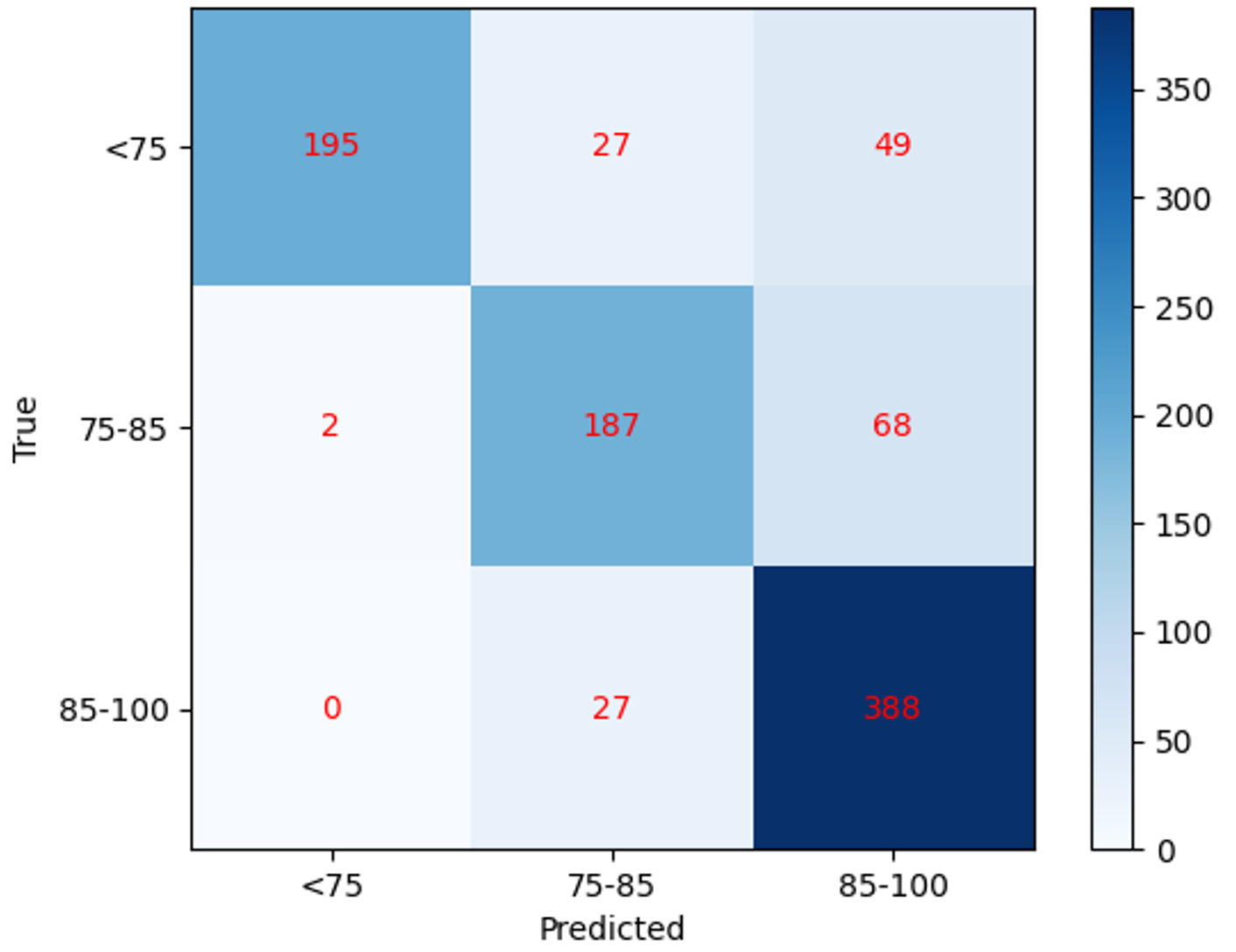}
    \caption{Confusion matrix of segment classification}
    \label{Fig: SVM classification}
\end{figure*}

During the fitting process, we use a segmental linear divided into four phases to integrate the four single evaluation terms, and a segment classifier based on SVM is applied to decide which level a driving event belongs to. The classification accuracy is illustrated in Fig.~\ref{Fig: SVM classification}. The overall accuracy is 81.65\%, and among the incorrect events, 86.71\% are classified into adjacent segments.

After implementing the segment classifier, the core hyperparameters that need fitting are the four corresponding weights $w_{safe}$, $w_{eff}$, $w_{comf}$, $w_{eng}$, and the offset $b$. The restrictions are that all weights should be non-negative. The score thresholds for the four stages are 70, 75, and 80, and for each segment a set of weights will be derived.

In the integrated fitting process, for each driving case we mark the average evaluation score of all raters as the ground truth, and the loss function is the mean absolute error to the ground truth score, as shown in ~\ref{MAE} where $S_{int}^{rev}$ is the calculated score and $S_{gt}$ is the ground truth. Since the original human evaluation ground truth is a score ranging from 0 to 100, the mean absolute error is also on a percentage scale. To decrease the impact of data distribution bias, for each set of parameters we randomly divide the dataset into train and validation set by 80\% and 20\%, ensuring that the validation data are not used for fitting weights. The final result is the average value of five experiments.

\begin{equation}
\label{MAE}
    MAE = \frac{1}{n} \sum|S_{int}^{rev} - S_{gt}|
\end{equation}

As introduced in section~\ref{3.C}, normalization is a crucial step in the model integration process, and we conducted ablation experiments to compare the performance differences of different normalization methods. The results are illustrated in Table.~\ref{Tb:normalization}, and the validation MAE of the listed methods are 4.58, 5.76, 6.46, and 7.59 out of 100, respectively. The max-min linear has the lowest MAE. Other methods such as z-score may introduce blurring in the original data distribution, thereby causing a loss of some inherent data characteristics and negatively affecting the segment classifier's performance. Hence, although other normalization methods have similar performance on linear model fitting, the errors resulting from inaccurate classification render max-min linear as the optimal choice.

\begin{table}[htb]
\footnotesize
	\begin{center}
		\caption{Ablation experiment of normalization methods}\label{Tb:normalization}
		\begin{threeparttable}
		\setlength{\tabcolsep}{5mm}{
		\begin{tabular}{cccccccc}
		\toprule
			Normalization method & Validation MAE(\%) \\\hline
			     max-min linear & 4.58 \\
                arctan & 5.76 \\
                robust scaling & 6.46 \\
                z-score & 7.59 \\
			\bottomrule
		\end{tabular}}
		\end{threeparttable}
	\end{center}
	\vspace{-5mm}
\end{table}

The final fitting results of the weights and offset are shown in Table.~\ref{Tb:fitting results}. For the safety term, the weights in low and medium are larger than that of high level, and the reason is that for driving events with scores higher than 85, safety is a prerequisite that all drivers share, thus making little difference on the final evaluation result. The efficiency term shows a similar trend, with a significant impact in the mid to low range, while the weight drastically decreases in the high score range. Conversely, comfort becomes a significant differentiating factor in the high score range, whereas its impact is limited in the mid to low ranges. Compared with the other three factors, energy consumption affects the overall evaluation evenly. In summary, for low-level driving cases, safety and efficiency become more decisive, and for high-level cases comfort becomes the dominating term. The fitting results conform to a human’s intuitive understanding, since safety and time efficiency are the most fundamental demands that decide whether the driving event is qualified, while comfort is a higher-level pursuit making a driving event pleasant. 

\begin{table}[htb]
\footnotesize
	\begin{center}
		\caption{Fitting results of weights and offset}\label{Tb:fitting results}
		\begin{threeparttable}
		\setlength{\tabcolsep}{5mm}{
		\begin{tabular}{cccccccc}
		\toprule
			Variable & Low level & Medium level & High level \\\hline
			     $w_{safe}$ & 0.165 & 0.160 & 0.010 \\
                $w_{eff}$ & 0.235 & 0.343 & 0.103 \\
                $w_{comf}$ & 0.010 & 0.161 & 0.507 \\
                $w_{eng}$ & 0.280 & 0.166 & 0.238 \\
                $b$ & \multicolumn{3}{c}{10} \\
			\bottomrule
		\end{tabular}}
		\end{threeparttable}
	\end{center}
	\vspace{-5mm}
\end{table}

After deriving the needed parameters, we compare our proposed algorithms to some baselines to validate the performance.

\subsection{Baselines}

For comparison, we select several previous decision-making planning methods as baselines, including single-dimension ones and comprehensive ones.
\subsubsection{Single factor evaluation baselines}
The first category of baselines are the four single-factor models designed in this research, which are safety, time efficiency, comfort, and energy consumption. The comparison between the single-factor methods and comprehensive ones will demonstrate whether considering multiple factors is essential for decision-making evaluation.

\subsubsection{Integrated intelligence evaluation}
Based on the Least Action Principle, this method merges risk, efficiency, comfort, and smoothness into one defined action quantity $S_{int}$ of ego vehicle $i$~\cite{huang2020integrated}, as shown in from~\eqref{IIE G} to ~\eqref{IIE}:

\begin{equation}
\label{IIE G}
    S_{int} = \int_{t_0}^{t_f} L_{int}dt
\end{equation}

\begin{equation}
\begin{split}
    \label{IIE}
    L_{int} = &-U_{risk} + U_{eff} + U_{com} + U_{smooth} \\
    = &\frac{1}{2}m_i[v_i^2 + \sum_{j=1}^{n-1}{(v_i - v_j)}^2- {(j_{ix} + j_{iy})}^2] \\
    &-\int_{t_0}^{t_f}[(R_i - G_i)v_{ix} + (F_{li,2} - F_{li,1})v_{iy} + \sum_{j=1}^{n-1}2F_{ji}(v_i-v_j)]dt
\end{split}
\end{equation}

\noindent where $t_0$ and $t_f$ are start and finish time of the driving case, $m_i$ is vehicle mass, $v_i$ is the velocity, $j_{ix}$ and $j_{iy}$ are longitudinal and lateral jerk, $R_i$ is the restriction resistance, $G_i$ is the virtual gravity, $F_{li,2}$ and $F_{li,1}$ are the lateral binding force of lane lines, and $F_{ji}$ is the external force caused by road user $j$ ~\cite{huang2020integrated}.

After defining the action quantity $S_{int}$, this method labels the average $S_{int}$ value of excellent drivers as the pole score, denoted as $S_{int}^*$, and normalizes the distribution of other driving case’s $S_{int}$ to a range between 0 and 1 linearly. The final position in the [0,1] range is the final evaluation result. In this study we also transfer it into a percentage scale.

\subsubsection{CRITIC-AHP-GRA}
This baseline divides driving evaluation into four major categories: safety, comfort, intelligence, and efficiency. Then, each category is further broken down into a linear combination of several physical quantities. Subsequently, the weights between physical quantities within each category are determined using the Analytic Hierarchy Process (AHP), followed by calculating the weights between categories using the criteria importance through intercriteria correlation (CRITIC) algorithm~\cite{zhao2022objective}. Through these two procedures, the needed relative weights are derived and confirmed. In this experiment, we use the weights recorded in the paper.

Similar to the second baseline, this method also selects the best single categories from human-driven cases as the pole scores and utilizes the Grey Relational Analysis (GRA) to output a similarity of the tested trajectory terms to the pole scores. Then combined with the previously derived weights, a final evaluation score is given.

\subsubsection{Penalty-based evaluation}

Currently many AD planning challenges use a penalty-based evaluation system, and we select the Onsite Competition’s published judging rules as a baseline in this experiment. This system considers three categories of driving: safety, efficiency, and comfort, with a point ratio of 50 to 30 to 20. Each aspect involves many penalty conditions, and if one condition is met in one timestep, a penalty will be given. At the end of the entire driving event, the final score is obtained by subtracting the penalty points from all-time steps from the total score. The detailed penalty conditions are illustrated in Table ~\ref{Tb:onsite}.

\begin{table}[htb]
\footnotesize
	\begin{center}
		\caption{Penalty conditions of Onsite evaluation system}\label{Tb:onsite}
		\begin{threeparttable}
		\setlength{\tabcolsep}{5mm}{
		\begin{tabular}{cccccccc}
		\toprule
			Category & Penalty term \\\hline
			     \multirow{5}{*}{Safety-50 points} & Crash \\
                                                    & Tight TTC\\
                                                    & Out of road boundary\\
                                                    & Wrong direction\\
                                                    & Run a red light\\\hline
                \multirow{2}{*}{Efficiency-30 points} & Not finish required route\\
                                                        & Too long driving time\\\hline
                \multirow{5}{*}{Comfort-20 points} & Large longitudinal velocity vibration\\
                                                    & Large longitudinal acceleration vibration\\
                                                    & Large lateral velocity vibration\\
                                                    & Large lateral acceleration vibration\\
                                                    & Large yaw rate vibration\\
			\bottomrule
		\end{tabular}}
		\end{threeparttable}
	\end{center}
	\vspace{-5mm}
\end{table}

\subsection{Comparison and Analysis}
\label{section analysis}
To compare the evaluation performance of the proposed method, we randomly generate a validation set five times, the amount of which is 20\% of the overall driving cases. The results are shown in Table~\ref{Tb:Main results}.

\begin{table}[htb]
\footnotesize
	\begin{center}
		\caption{Evaluation results}\label{Tb:Main results}
		\begin{threeparttable}
		\setlength{\tabcolsep}{5mm}{
		\begin{tabular}{cccccccc}
		\toprule
			Category & Method & Validation MAE \\\hline
			     \multirow{4}{*}{Single factor evaluation} & Safety & 11.52 \\
                                                    & Efficiency & 9.38\\
                                                    & Comfort & 10.50\\
                                                    & Energy & 8.90\\
                                                    & Run a red light\\\hline
                \multirow{5}{*}{Comprehensive evaluation} & Integrated intelligence evaluation & 6.79\\
                                                        & CRITIC-AHP-GRA & 8.40\\
                                                        & Penalty-based evaluation & 11.20 \\
                                                        & Linear fitting & 6.36 \\
                                                        & S2O & 4.58 \\
			\bottomrule
		\end{tabular}}
		\end{threeparttable}
	\end{center}
	\vspace{-5mm}
\end{table}

The validation MAE of our proposed method is 4.58, and the baselines’ MAE ranges from 6.58 to 11.52. Compared with the best baseline, our method reduces the evaluation error by 32.55\%. It can be observed that a comprehensive evaluation system is overall far better than the single factor methods, which conforms to human’s intuition and cognition of evaluation. Driving decision evaluation is a complicated process influenced by multiple factors, and single-dimensional tests cannot prove that an AD algorithm is better.

Compared with the comprehensive models, although the gap is much smaller, our method still obviously outperforms. The Integrated intelligence evaluation (IIE) suffers from the inaccurate modelling of single terms, such as representing the risk and efficiency only by combined calculation of speed metrics. In contrast, our method used revised single-term models that combine the related works, reflecting driver experience more realistically. On the other hand, although CRITIC-AHP-GRA (CAG) covers a variety of physical quantities, one critical problem is that factor analysis can only capture the correlation among variables, which does not necessarily equate to their contribution to the final comprehensive evaluation. Thus, directly using the correlation coefficients as weights is biased. Compared with CAG, our method uses the subjective rater score to solve the objective weights, skipping the intermediary process. As for the PE baseline, the pivotal problem is that the penalty mechanism is not appropriate for evaluation. For example, when a vehicle is involved in a collision, only the applicable moments that meet the penalty criteria will trigger a deduction of points, which is often for a short period of time. However, the preceding actions are also causes of accidents and should not be overlooked. This issue results in the system still giving tens of points after a crash, deviating significantly from the ground truth. In contrast, our model considers the mean term value during the whole driving event, expressing a continuous evaluation result.

Another result worth mentioning is that our method shows little generalization error, with a gap of only 0.02 between the train error and validation error. The reason is that our proposed model is fundamentally a rule-based method and does not employ black-box deep learning approaches, thus avoiding overfitting to the data and being interpretable. This result ensures a good generalization ability in real-world applications.

From the perspective of human factors research, a mean absolute error of 4.58 is a precise evaluation level. Firstly, human ratings inherently have a range of variability. The same individual person may exhibit a deviation of approximately 5 points even when rating the same driving case. Therefore, completely reducing the error to a near-zero level is nearly impossible to achieve.

\subsection{Feasibility verification}
\subsubsection{Embedment on real-time simulation platform}
\begin{figure*}
    \centering
    \includegraphics[width=10cm]{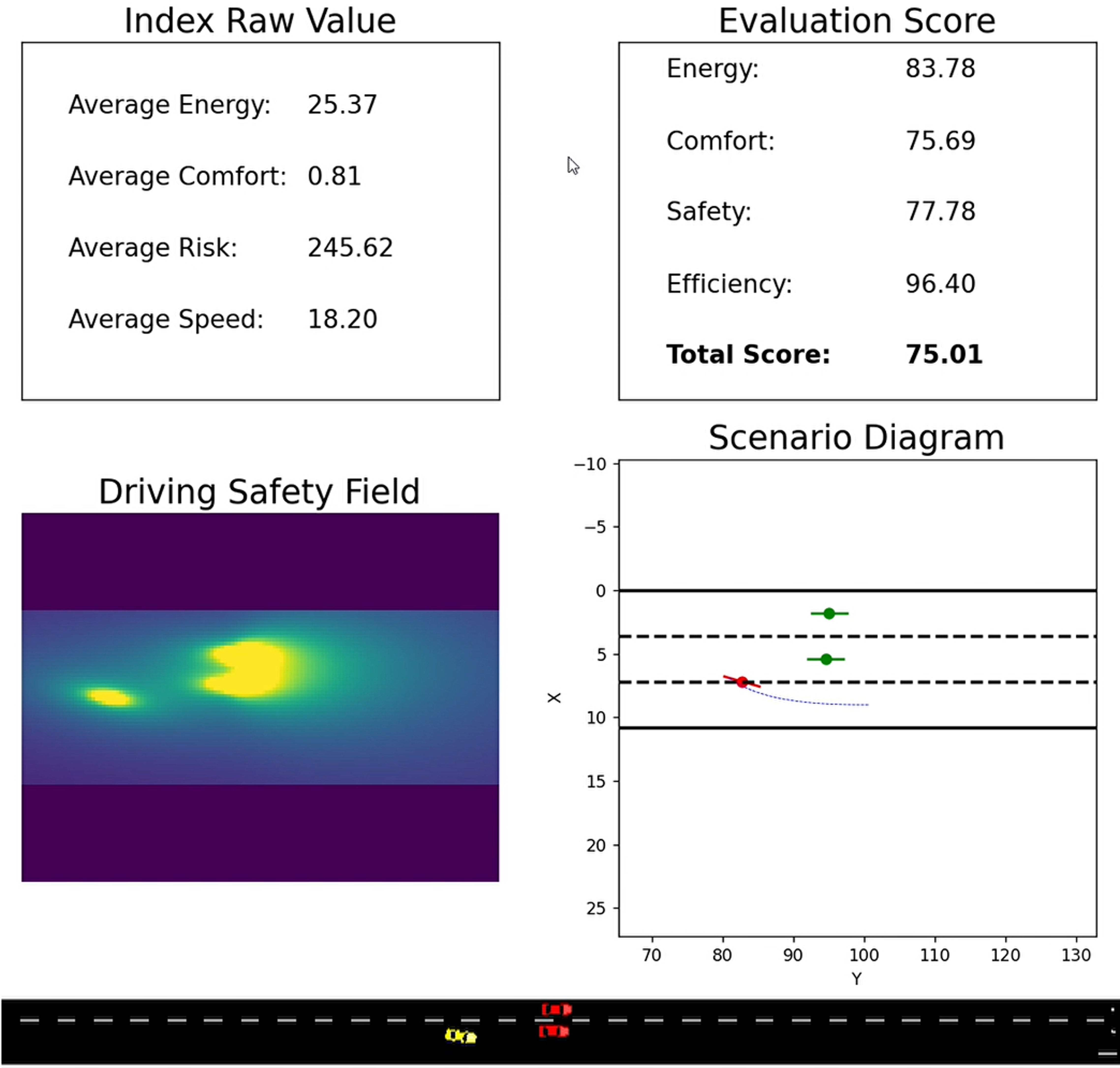}
    \caption{Visualization interface of online decision-making evaluation on SUMO and the original GUI interface of SUMO.}
    \label{Fig: SUMO visual}
\end{figure*}

To verify the feasibility of our proposed method, we embed the evaluation algorithm into SUMO, a simulation platform widely used in transportation and vehicle-related research, to conduct online driving decision-making evaluation, as shown in Fig.~\ref{Fig: SUMO visual} which shows the visualization interface and the original SUMO GUI.

In this example, we run a default decision-making algorithm within SUMO. The physical states of agents and the roadmap are continuously updated and transferred into the evaluation module, and assessment results are computed in real time. The visualization interface, which is divided into four sections, can output real-time scores for the four individual factor scores, the total score, the Driving safety field heat map that is used for safety evaluation, and the diagram of the traffic scenario. The implementation of S2O on the SUMO platform demonstrates the feasibility of our algorithm.

\subsubsection{Validation of decision-making algorithm performance evaluation}
To directly validate the precision and correctness of the decision-making evaluation function, based on the above real-time platform we test the driving performance of three different autonomous driving decision methods and compare with the alignment of human subjective evaluations.

\begin{figure*}
    \centering
    \includegraphics[width=14cm]{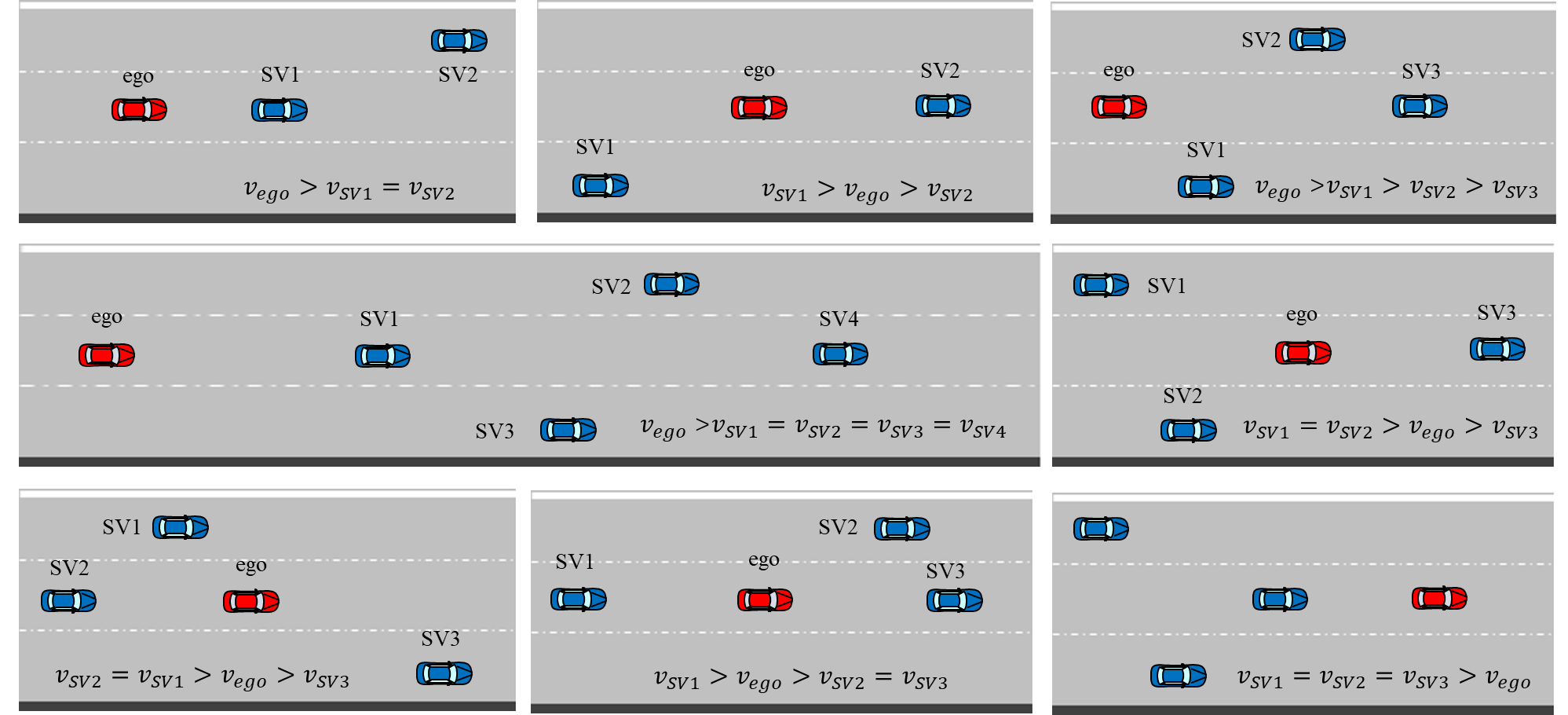}
    \caption{Test scenarios of decision-making performance evaluation.}
    \label{Fig: test scene}
\end{figure*}

The three decision-making methods used in this experiment are Intelligent Driver Model (IDM)~\cite{kreutz2021analysis} and two optimization-based models with different weights on driving factors~\cite{wang2024homogenuous, eiras2021two}. In the validation, we run the three tested AD planning algorithms on eight interactive scenarios which could generate differentiated behaviors, as shown in Fig.~\ref{Fig: test scene}. In these test scenarios, the ego vehicle needs to driving forward facing the interactions from surrounding vehicles (SV) ranging from 2 to 4. To maintain high passing efficiency and safety level, the ego vehicle may conduct complicated behaviors such as overtaking preceding slow SV and yielding to following quick SV.

Similar to the collection of D2E dataset, in the validation phase we recruit 10 third-party volunteers with diverse distributions to give a subjective evaluation score on each of the driving events and analyze whether the results derived from S2O aligns with human feelings. The results are shown in Tabel.~\ref{Tb:validation}, where $S_h$ denotes the average human subjective evaluation score, $S_{S2O}$ denotes the score caluated from the proposed S2O model, $MAE$ denotes the deviation from $S_{S2O}$ to $S_h$, $R_h$ denotes the rank of the algorithm according to human evaluation, and $R_{S2O}$ denotes the rank according to S2O. 

\begin{table}[htb]
\footnotesize
	\begin{center}
		\caption{Validation of decision-making algorithm performance evaluation}\label{Tb:validation}
		\begin{threeparttable}
		\setlength{\tabcolsep}{1.8mm}{
		\begin{tabular}{c|ccccc|ccccc|ccccc}
		\toprule 
			\multirow{2}{*}{Scene ID} & \multicolumn{5}{c|}{Algorithm A} & \multicolumn{5}{c|}{Algorithm B} & \multicolumn{5}{c}{Algorithm C}\\
                                 & $S_h$ & $S_{S2O}$ & $MAE$ & $R_h$ & $R_{S2O}$  & $S_h$ & $S_{S2O}$ & $MAE$ & $R_h$ & $R_{S2O}$ & $S_h$ & $S_{S2O}$ & $MAE$ & $R_h$ & $R_{S2O}$\\\hline
			     1 & 86.4 & 81.42 & 4.96 & 2 & 2 & 90.3 & 82.39 & 7.86 & 1 & 1 & 74.4 & 79.34 & 3.97 & 3 & 3 \\
                 2 & 82.5 & 80.64 & 1.86 & 1 & 1 & 81.5 & 78.21 & 3.29 & 2 & 3 & 77.0 & 79.22 & 2.22 & 3 & 2\\
                 3 & 91.6 & 82.21 & 9.42 & 1 & 1 & 76.9 & 78.21 & 1.33 & 3 & 3 & 79.4 & 79.40 & 0.03 & 2 & 2\\
                 4 & 86.1 & 81.46 & 4.67 & 2 & 1 & 86.5 & 81.41 & 5.09 & 1 & 2 & 80.9 & 79.85 & 1.03 & 3 & 3\\
                 5 & 73.5 & 72.47 & 1.03 & 2 & 2 & 69.3 & 65.49 & 3.76 & 3 & 3 & 80.8 & 80.58 & 0.17 & 1 & 1\\
                 6 & 76.4 & 77.51 & 1.14 & 3 & 2 & 77.4 & 74.50 & 2.88 & 2 & 3 & 82.5 & 83.92 & 1.42 & 1 & 1\\
                 7 & 86.3 & 79.62 & 6.63 & 1 & 2 & 78.9 & 72.35 & 6.53 & 3 & 3 & 81.1 & 85.59 & 4.47 & 2 & 1\\
                 8 & 77.5 & 75.22 & 2.28 & 3 & 2 & 83.5 & 73.66 & 9.84 & 1 & 3 & 80.5 & 85.21 & 4.71 & 2 & 1\\
			\bottomrule
		\end{tabular}}
		\end{threeparttable}
	\end{center}
	\vspace{-5mm}
\end{table}

According to the validation results, the proposed S2O model can evaluate autonomous driving decision-making algorithms precisely, with a mean absolute error of 3.77 under a percentage scale. Note that the MAE in the validation phase is less than the results on D2E dataset, because the tested AD algorithms all ensure a basic acceptable performance and scores of most cases range from 75 to 90. As discussed in Section \ref{section analysis}, S2O has better evaluation precision on regular cases than extreme ones. From the perspective of ranking, the average ranking error is 0.50 compared with human subjective results. Although S2O evaluate most of the cases accurately, the evaluation of Scene 8 suffers from relatively large MAE. The reason is that Scene 8 is a complicated situation where all three fast vehicle from behind try to overtake the ego vehicle, which is not common in daily driving experience. Therefore, various raters have difference comments on the driving performance. This phenomenon demonstrates that performance evaluation of extremely complex scenarios is still a challenge for both human and objective models. Nevertheless, overall S2O can effectively evaluate AD algorithm performances in a real-time and comprehensive view.

\section{Conclusions and Future Work}
\label{Sec:5}

This paper proposed S2O, an integrated driving decision-making performance evaluation method bridging subjective human assessment to objective evaluation. Aimed at the challenge of non-comprehensive consideration of various driving experience factors and over-simplistic and intuitive integration models, our method establishes modified fundamental models of safety, time efficiency, comfort, and energy efficiency, and designs a segment linear fitting in conjunction with an SVM segment classifier based on human rating distribution analysis. According to experiment results, the proposed S2O method achieves a mean absolute error of 4.58 to human evaluation ground truth under a percentage scale. Compared with baselines, the error is reduced by 32.55\%, increasing evaluation accuracy while maintaining interpretability. Online implementation of S2O on SUMO platform and real-time validations on the performance evaluation of three autonomous driving decision-making algorithms prove the feasibility and effectiveness.

Future work will be devoted to exploring a more personalized evaluation mechanism, realizing a balance between common regularities and personal demands. In addition, the evaluation of extreme driving cases is still an open challenge, and we plan to further optimize the proposed S2O model. We also want to develop S2O into an open-source website leaderboard so that autonomous driving developers can easily evaluate and compare the performance of their planning algorithms with others.

\section*{Acknowledgement}
This work is supported by the National Natural Science Foundation of China under Grant 52131021 (the key project), and 52221005. This work is also supported by the Joint Laboratory for Internet of Vehicles, Ministry of Education-China Mobile Communications Corporation.

\bibliography{mybibfile}

\end{document}